\documentclass{article}

     \usepackage[preprint]{neurips_2020}

\usepackage[utf8]{inputenc} 
\usepackage[T1]{fontenc}    
\usepackage{hyperref}       
\usepackage{url}            
\usepackage{booktabs}       
\usepackage{amsfonts}       
\usepackage{nicefrac}       
\usepackage{microtype}      

\usepackage{color}
\usepackage{multirow}
\usepackage{xspace}
\usepackage{graphicx}
\usepackage{comment}
\usepackage{amsmath,amssymb}
\usepackage{subfigure}
\usepackage{algorithmic}
\usepackage{algorithm}
\usepackage{enumitem}
\newcommand{\irnas}{DI-NAS\xspace}

\def\ie{\emph{i.e.,~}}
\def\eg{\emph{e.g.,~}}

\def\etc{\emph{etc.}}
\def\wrt{\emph{w.r.t.~}}

\def\yifan{\textcolor{black}}

\definecolor{chenyaofo}{RGB}{17,120,100}

\def\ljc{\textcolor{black}} \def\guo{\textcolor{black}}

  \def\mD{{\mathcal D}}

  \def\mL{{\mathcal L}}

  \def\mP{{\mathcal P}}
  \def\mQ{{\mathcal Q}}
  \def\mR{{\mathcal R}}

  \def\mX{{\mathcal X}}

  \DeclareMathAlphabet\mathbfcal{OMS}{cmsy}{b}{n}

  \def\0{{\bf 0}}
  \def\1{{\bf 1}}

  \def\bG{{\bf G}}
  
  \def\bI{{\bf I}}

  \def\bP{{\bf P}}

  \def\bW{{\bf W}}
  \def\bX{{\bf X}}

  \def\bx{{\bf x}}

  \def\mmR{{\mathbb R}}

  \def\mmF{{\mathrm F}}

  \def\bx{{\bf x}}
  \def\bX{{\bf X}}

  \def\bW{{\bf W}}

  \def\bP{{\bf P}}

  \def\citep{\cite}

  \def\yifan{\textcolor{black}}

\newtheorem{theorem}{Theorem}[section]

\title{Disturbance-immune Weight Sharing for \\  Neural Architecture Search}

\author{
    Shuaicheng Niu\thanks{Authors contributed equally.}~, Jiaxiang Wu$^*$, Yifan Zhang$^*$,  \\ \textbf{Yong Guo, Peilin Zhao, Junzhou Huang, Mingkui Tan\thanks{Corresponding author.}} \\
    South China University of Technology, Tencent AI Lab, \\
    University of Texas at Arlington\\
    \{sensc, sezyifan, guo.yong\}@mail.scut.edu.cn, \{mingkuitan\}@scut.edu.cn, \\
     \{jonathanwu, masonzhao\}@tencent.com, jzhuang@uta.edu
}

\begin{document}

\maketitle

\begin{abstract}
  Neural architecture search (NAS) has gained increasing attention in the community of architecture design. One of the key factors behind the success lies in the training efficiency created by the weight sharing (WS) technique. However, WS-based NAS methods often suffer from a performance disturbance (PD) issue. That is, the training of subsequent architectures inevitably disturbs the performance of previously trained architectures due to the partially shared weights. This leads to inaccurate performance estimation for the previous architectures, which makes it hard to learn a good search strategy. To alleviate the performance disturbance issue, we propose a new disturbance-immune update strategy for model updating. Specifically, to preserve the knowledge learned by previous architectures, we constrain the training of subsequent architectures in an orthogonal space via  orthogonal gradient descent. Equipped with this strategy, we propose a novel disturbance-immune training scheme for NAS. We theoretically analyze the effectiveness of our strategy in alleviating the PD risk. Extensive experiments on CIFAR-10 and ImageNet verify the superiority of our method.
\end{abstract}

\section{Introduction}
Deep neural networks (DNNs) have produced state-of-the-art results in many challenging tasks, such as image classification~\cite{resnet,guo2018double}, face recognition~\cite{SphereFace,schroff2015facenet}, medical image analysis~\cite{zhang2019collaborative, zhang2019whole}, portfolio selection~\cite{zhang2018adaptive}, and image generation~\cite{cao2019multi},  \etc~One of the key factors behind the success of DNNs lies in the design of effective neural architectures, such as ResNet~\cite{resnet} and MobileNet~\cite{howard2017mobilenets}.
In practice, different architectures often show different performances, and the optimal architecture may vary among different tasks.
Hence, the design of effective neural architectures highly relies on substantial human expertise. However, the human-designed process cannot fully explore the whole architecture space, resulting in suboptimal architectures~\cite{zoph2016neural}.

Besides manual design, one may resort to the neural architecture search (NAS)~\cite{zoph2016neural} technique to automatically design network architectures.
Specifically, NAS seeks to find an optimal architecture in a predefined search space~\cite{zoph2016neural,zoph2018learning}.
The searched architectures often show promising performance, demonstrating tremendous potential to surpass hand-crafted architectures~\cite{cai2018efficient,cai2018proxylessnas,liu2018progressive,liu2017hierarchical,tan2019mnasnet}.
During the search process, NAS seeks to evaluate a large number of candidate architectures and gradually learns a strategy to find good architectures.
However, the exact performance evaluation requires training all the architectures from scratch, which is highly time-consuming and computationally impractical in real-world applications.

To improve the training efficiency, a weight sharing (WS)~\cite{bender2018understanding,pham2018efficient} technique has been developed for NAS. Specifically, WS-based NAS constructs a supernet,
\ie a large computational graph, where each network architecture (subgraph) shares their parameters.
In this sense, all candidate architectures share its parameters in the supernet. Based on the supernet, one can directly estimate the performance of architectures instead of training them from scratch. In this way, WS is able to effectively accelerate the search process of NAS and reduce the search cost from 1,800 GPU days to less than 1 day~\cite{cai2018proxylessnas}. Despite the efficiency, recent studies~\cite{adam2019understanding,li2019random,sciuto2019evaluating} have empirically shown that the performance estimation provided by the WS is inaccurate, which makes it hard to identify good architectures.
As a result, the search performance of NAS cannot be guaranteed and often fails to find promising architectures~\cite{chu2019fairnas,luo2019understanding}.

In this paper, we study the risk of the WS scheme and find that WS-based methods suffer from a performance disturbance (PD) issue. That is, the training of a subsequent architecture inevitably disturbs the performance of previously trained architectures due to the update of shared parameters. As a result, the performance estimation of the previous architectures is unstable and inaccurate, which makes it difficult to search for good architectures. To address this issue, we propose a new disturbance-immune update strategy to train the WS. By exploring orthogonal gradient descent, the proposed strategy trains architectures for better performance while constraining the prediction of previously trained architectures to be unchanged. In this way, we are able to alleviate the PD issue in WS and provide more stable and accurate performance estimation. Based on this update strategy, we further propose a novel disturbance-immune training scheme for NAS, which helps to search for better architectures.

The main contributions of this paper are summarized as follows.

\begin{itemize}[leftmargin=*]
    \item We propose a novel disturbance-immune WS training scheme for NAS. By updating models in an orthogonal space with orthogonal gradient decent, our method exhibits more stable/accurate performance estimation for NAS. Equipped with the disturbance-immune WS,  NAS is able to learn a better search strategy and find better architectures.

    \item  We theoretically verify the effectiveness of the proposed disturbance-immune training scheme in alleviating the performance disturbance risk of WS. We also provide an asymptotic convergence analysis of the proposed method.

    \item Extensive experiments demonstrate that the proposed method is able to alleviate the performance disturbance issue and provide more accurate performance evaluation for strategy learning. As a result, the architecture searched by our method performs better than the architectures obtained by other state-of-the-art NAS methods on the CIFAR-10 and ImageNet datasets.
\end{itemize}

\section{Related Work}

\textbf{Neural architecture search.}
In the past few years, neural architecture search (NAS) has attracted increasing attention to automatically design effective architectures.
\ljc{The classical NAS problem \cite{zoph2016neural} exploits the paradigms of reinforcement learning (RL) to generate the model descriptions of DNNs.}
\ljc{After that, MetaQNN \cite{baker2016designing} automatically selects the architectures of DNNs through a RL-based meta-modeling procedure. NASNet \cite{zoph2018learning} designs a new search space to improve the search performance.}
\ljc{Moreover, some studies \cite{real2018regularized,liu2017hierarchical} use evolutionary algorithms to find new architectures with excellent performance.}
To guide the search process, the NAS methods need to estimate the performance of candidate architectures.
The simplest way is to train candidate architectures from scratch to obtain the performance, which, however, is time-consuming and computationally expensive (\eg several thousands of GPU days).
To reduce the computational resources, recent NAS methods adopt \ljc{a} weight sharing \cite{pham2018efficient,liu2018darts,cai2018efficient,wu2019fbnet} strategy to estimate the performance of candidate architectures.

\textbf{Weight sharing approaches.}
Efficient neural architecture search (ENAS)~\cite{pham2018efficient} \ljc{first proposes} a NAS training scheme with weight sharing (WS), which measures the performance of an architecture with the weights inherited from the trained supernet.
Since WS can reduce the computational resources from thousands of GPU days to one \ljc{GPU day}, it is widely adopted to exploit NAS \ljc{in} various applications, such as objection detection~\cite{ghiasi2019fpn,chen2019detnas},  segmentation~\cite{liu2019auto} and compact architecture design~\cite{cai2018efficient,wu2019fbnet,tan2019mnasnet}.
Besides, DARTS~\cite{liu2018darts} exploits the WS scheme with a continuous relaxation of the search space to search \ljc{for} promising architectures.
However, recent studies~\cite{li2019random,sciuto2019evaluating} find that the architecture performance measured by WS training is often very inaccurate, thus leading to the inferior performance of WS based NAS methods.
To address this, NAO-V2~\cite{luo2019understanding} improves WS based NAS by training candidate architectures adequately and training complex architectures more.
FairNAS~\cite{chu2019fairnas} proposes a fair WS training strategy, which can be only applied to a single-path search space instead of a cell-based search space.
Unlike existing methods, we identify a performance disturbance (PD) issue that occurred in WS training, and propose to achieve more accurate performance estimations by alleviating this PD issue.

\textbf{Continual learning.}
Continual learning (CL) aims to continuously learn a series of tasks~\cite{kirkpatrick2017overcoming,zeng2019continual,finn2017model,farajtabar2019orthogonal}.
One of the major issues in CL is catastrophic disturbance, \ie deep models often forget about previous tasks when model weights are updated for a new task.
To \ljc{address this}, EWC~\cite{kirkpatrick2017overcoming} attempts to impose constraints on the updating of model weights based on measuring the importance of previous tasks.
Moreover, regularization based methods ~\cite{he2017overcoming,zeng2019continual} neglect gradients for new tasks.
\ljc{In this paper, motivated by CL, we propose a novel disturbance-immune update strategy to effectively improve the performance of WS-based NAS methods.}

\section{Problem Definition and Motivation}

\noindent \textbf{Notations.}
Throughout the paper, we use the following notations.
Let $\Omega$ be the search space of the NAS. Given any architecture $ \alpha \in \Omega$, let $w(\alpha)$  be its trainable parameters and $w^*(\alpha)$ be its optimal model parameters trained on some datasets (\eg CIFAR-10 \& ImageNet). Moreover, let $||\cdot||_\mmF$ denote the Frobenius norm, and let $[n]=\{1,\dots,n\}$.

\subsection{Neural Architecture Search and Weight Sharing}

Neural architecture search (NAS) aims to \guo{search for} an optimal architecture \guo{from} a predefined search space.
In this paper, we focus on reinforcement learning (RL)-based methods~\cite{zoph2016neural,zoph2018learning}, which seek to learn a controller with parameters $\theta$ to generate architectures (\ie $\alpha\ {\sim} \pi(\alpha; \theta)$).
To \guo{find promising} architectures, these methods learn the controller by maximizing the expectation of \guo{architecture performance using} metric $\mathcal{R}(\alpha, w^*(\alpha))$ (\eg the accuracy on the validation set):
\begin{align}\label{eq:nas_theta}
    \max_{\theta}~\mathbb{E}_{\alpha\sim \pi(\alpha;\theta)}\mathcal{R}\left(\alpha, w^*(\alpha)\right), ~~\mathrm{s.t.}~ w^*(\alpha)=\arg\min_{w(\alpha)}\mathcal{L}\left(\alpha,w(\alpha)\right),
\end{align}
where $\mathcal{L}\left(\alpha,w(\alpha)\right)$ is the training loss on the training data. However, \guo{we have to} train candidate architectures from scratch to obtain $w^*(\alpha)$, resulting in an unbearable computational burden. To address this, a weight sharing scheme is proposed.

Weight sharing (WS) for NAS~\cite{pham2018efficient} constructs a supernet, \ie a large computational graph, where each network architecture (subgraph) shares their parameters.
Let $w$ be the \guo{parameters} of the  whole supernet and $w(\alpha)$ be the parameters of $\alpha$ inherited from the supernet.
\guo{To train the supernet, one can sample sufficiently many architectures $(\alpha_i)_{i\in[n]}$ and train them in a sequential manner~\cite{pham2018efficient}:}
\begin{align}\label{eq:ws_general}
    \min_{w(\alpha_i)}\mL (\alpha_i, w(\alpha_i)),~\forall i\in [n].
\end{align}
Despite the training efficiency, the \guo{performance estimated} by WS is often very inaccurate~\cite{adam2019understanding,li2019random,sciuto2019evaluating} due to the sequential training method of WS.
Specifically, the training of a subsequent architecture inevitably disturbs the performance of previously trained architectures. As a result, the \guo{estimated} performance of the previous architectures becomes inaccurate, \guo{making it hard to learn a good controller and search for good architectures.}

\begin{figure*}[!t]
\centering
\subfigure[Example of weight sharing (WS)]
{
	\label{fig:supernet}
	\includegraphics[width=0.48\linewidth]{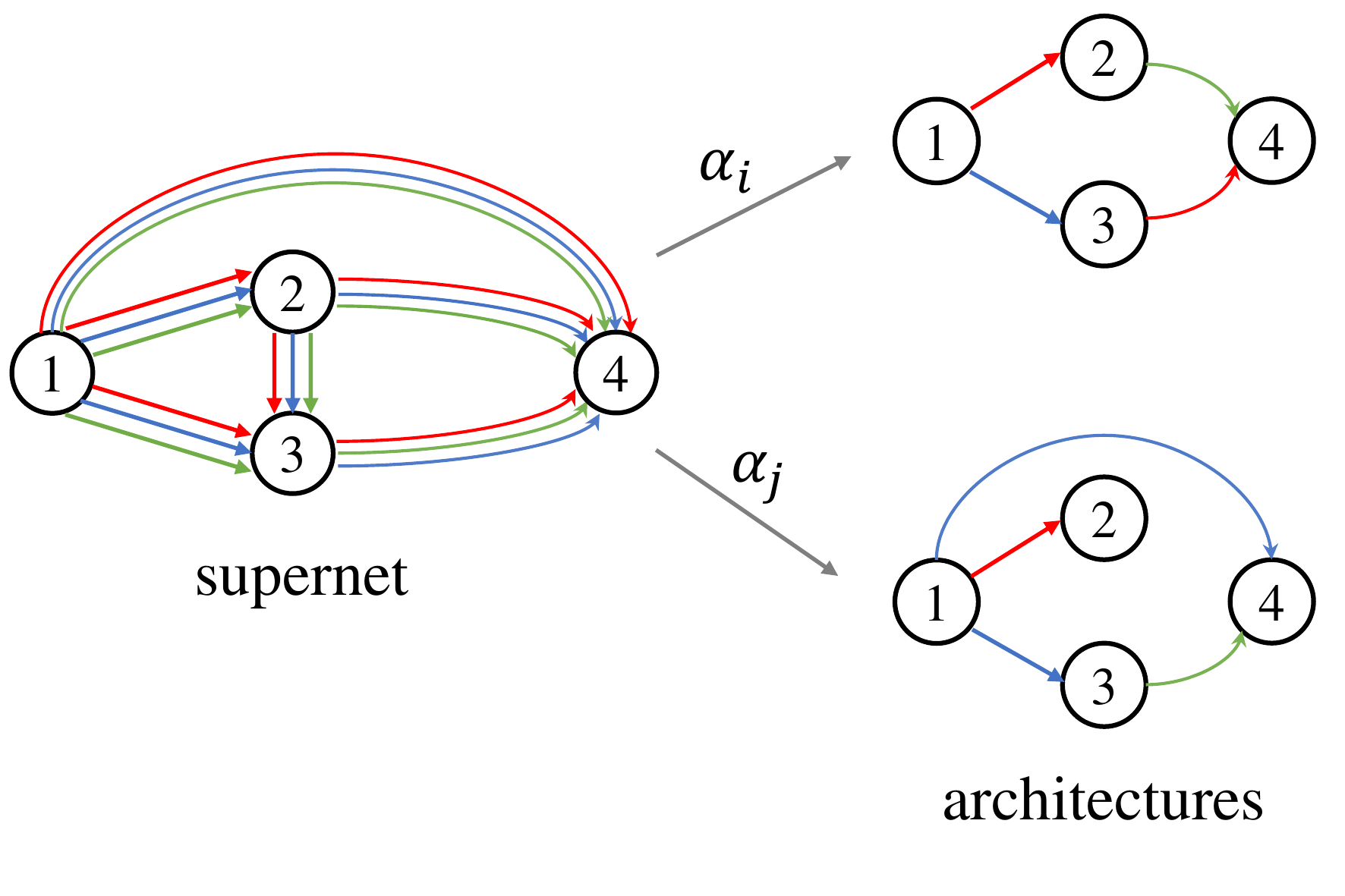}
}
\subfigure[Performance disturbance in WS]
{
	\label{fig:disturbance}
	\includegraphics[width=0.48\linewidth]{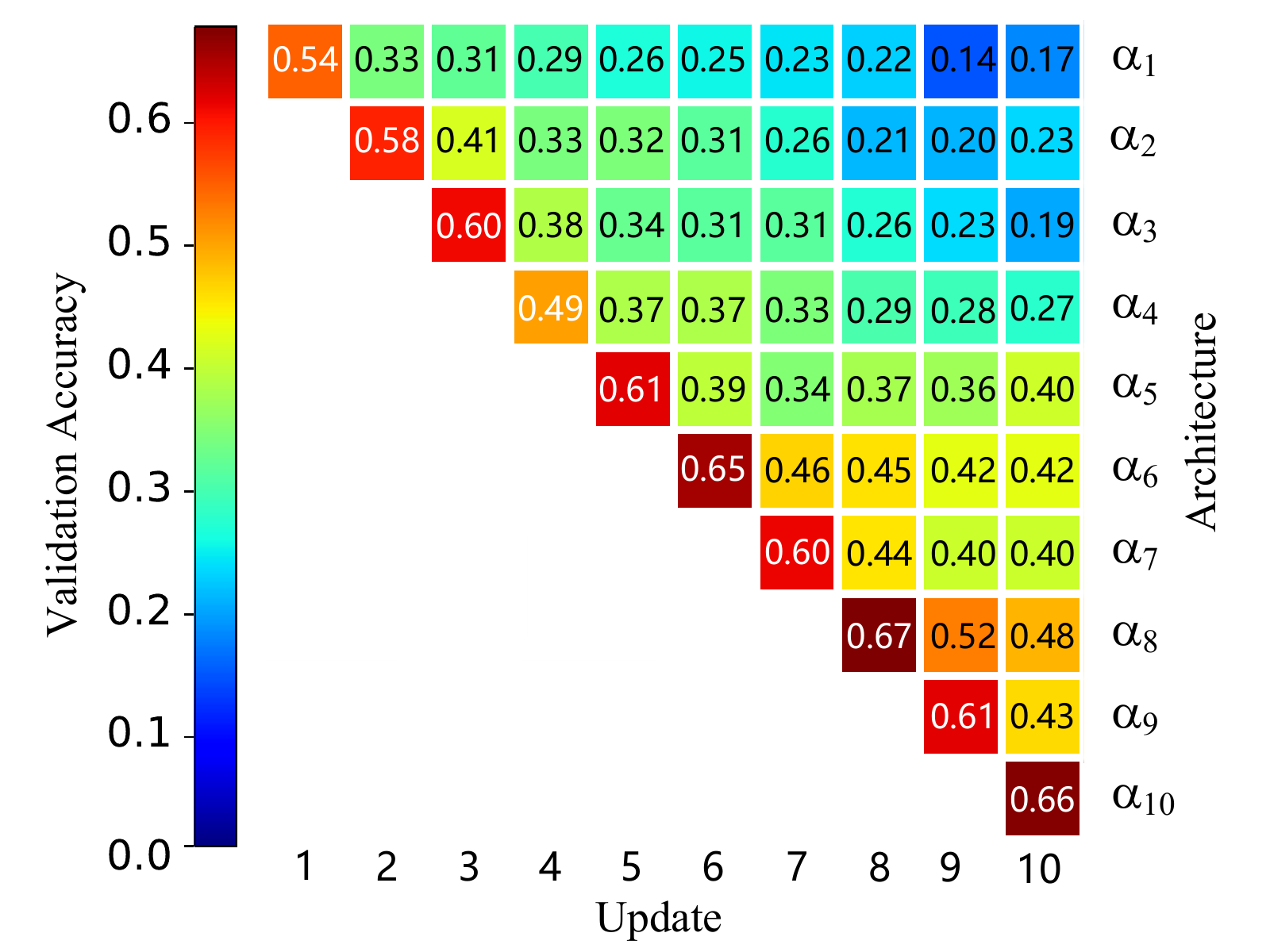}
}
\caption{Exemplar illustration of WS and performance disturbance (PD).
(a) The weights of architectures $\alpha_i$ and $\alpha_j$ are inherited from the supernet.
There are some shared weights in both $\alpha_i$ and $\alpha_j$.
(b) We record the performance of 10 architectures $(\alpha_i)_{i\in[10]}$. Element $(i,j)$ denotes the accuracy of $\alpha_i$ after the training of $\alpha_j$. Note that the training of the subsequent architectures results in apparent performance disturbance of the previously trained architectures.}
\label{fig:pi}
\vspace{0.1in}
\end{figure*}

\subsection{Performance Disturbance in Weight Sharing}

During the training of WS, for any two architectures $\alpha_i$ and $\alpha_j$ with $i {<} j$, there are some shared parameters that appear in both $\alpha_i$ and $\alpha_j$ (see examples in Fig.~\ref{fig:supernet}).
Due to the sequential training strategy in Eqn.~(\ref{eq:ws_general}), once we update the parameters of $\alpha_j$, the shared parameters would also be changed and $\alpha_i$ may yield different prediction results.
As a result, the architecture $\alpha_i$ may incur severe
\textbf{performance disturbance (PD)} and the reward $\mR(\alpha_i, w(\alpha_i))$ becomes inaccurate. To justify this, we show an illustrative example in Fig.~\ref{fig:disturbance}.

In Fig.~\ref{fig:disturbance}, we consider training 10 architectures $(\alpha_i)_{i\in[10]}$ sequentially with shared parameters on CIFAR-10.
We record their performance in terms of validation accuracy after the training of each architecture.
The element $(i,j)$ in Fig.~\ref{fig:disturbance} denotes the validation accuracy of $\alpha_i$ after the training of $\alpha_j$.
The $i$-th row illustrates the performance disturbance of $\alpha_i$ during the whole training process.
Moreover, the more subsequent architectures are trained, the more severe performance disturbance is.
Such unstable and inaccurate performance estimation would mislead learning the policy.

The performance disturbance can be measured by the predictions of an architecture, since the predictions directly determine architecture performance.
Let $\bX$ be the input data and $f(\bX; w(\alpha))$ be the prediction of $\alpha$ based on its parameters $w(\alpha)$.
For convenience,
we use $w_o(\alpha)$ and $w(\alpha)$ to represent the original parameters and the latest parameters after the training of other architectures that have shared parameters, respectively. During the training of WS, the performance disturbance of $\alpha$ can be measured by the differences of predictions:
\begin{equation}\label{eq:predict_difference}
    {\rm PD}(\bX, \alpha) := || f(\bX;  w(\alpha)) - f(\bX; w_o(\alpha)) ||_\mathrm{F}.
\end{equation}

\subsection{Disturbance-immune Weight Sharing Problem}\label{sec:motivation}

To reduce the prediction differences, we seek to define and address a disturbance-immune problem. Similar to model compression methods~\cite{luo2017thinet,zhuang2018discrimination}, one possible way is to restrict the changes of feature maps in each layer. In this sense, if we can update the shared parameters while maintaining the same feature maps for all previously trained architectures, the performance of these architectures will not be changed.
Specifically, for the $\kappa$-th update, we seek to improve the performance of $\alpha_\kappa$ and reduce the performance disturbance of all the previous architectures $(\alpha_i)_{i\in[\kappa-1]}$.
In this way, the weight sharing scheme is able to provide a more stable and accurate reward (or performance estimation) for all candidate architectures.

To this end, we propose a disturbance-immune training strategy that makes the update term of the model parameters orthogonal to the input features.
Note that all the layers adopt the same parameter update method. For simplicity, we only investigate the training process \wrt a single layer.
Specifically, during the training of the WS, for any layer of the supernet, we record the input feature maps $\bX$ of all the previously trained architectures in $\mX$.
To ensure that we can improve performance \wrt the current architecture, we do not include its input feature in $\mX$.
In order to reduce PD of the previously trained architectures, we make the gradient \wrt the shared parameters orthogonal to all the input feature maps in $\mX$. In this way, there would be no change in the prediction results for the previous architectures.
Formally, let $\bW$ be the parameters of a specific layer and $\bW_o$ be the original parameters before the model update. The parameters $\bW$ have to satisfy a constraint that restricts the changes of output feature maps and forms a feasible set
\begin{equation}
\mQ=\{\bW: ||\bW^\top\bX-\bW_o^{\top}\bX||_\mathrm{F}\leq\gamma, \forall~\bX \in \mX \},
\end{equation}
where $\gamma$ is some small positive value. When we apply such constraints to train WS, the optimization problem becomes:
\begin{equation}
    \min_{w(\alpha_i)}\mL (\alpha_i, w(\alpha_i)), ~~\forall i\in[n]
    ~~\mathrm{s.t.}~\bW\in \mQ.
    \label{eq:new_constraint}
\end{equation}

Unlike the original problem in Eqn.~(\ref{eq:ws_general}), we seek to train the WS by keeping the output features of all the previously trained architectures unchanged.
In this sense, the performance of these architectures becomes stable and the PD of them can be greatly reduced (see results in Fig.~\ref{fig:hpnas_epoch_interval_13}).
Note that recording all the input features is infeasible in practice. To address this issue, we propose an equivalent solution that adopts the iterative training scheme and avoids data recording (see details in Section~\ref{sec:gradient}).
More critically, we theoretically prove that the proposed method is able to reduce the risk of performance disturbance (see theoretical analysis in Section~\ref{sec:theory}). We call this method Disturbance-immune Weight Sharing and show the details in Section~\ref{sec:dfws}.

\section{Disturbance-immune Weight Sharing}\label{sec:dfws}

In this section, we first propose a  disturbance-immune update strategy for architecture training with weight sharing (WS). Such a strategy aims to handle performance disturbance (PD) by constraining the change of feature maps of previously trained architectures.
Based on this strategy, we propose a new disturbance-immune training scheme for neural architecture search (\irnas).
Lastly, we theoretically analyze the effectiveness of the proposed method.


\subsection{Disturbance-immune Update Strategy}\label{sec:gradient}
To address PD, we propose to project the update gradient towards a useful direction but avoid large changes regarding the feature maps of shared parameters. To this end, we resort to the orthogonal gradient descent method~\cite{zeng2019continual}. Such a method projects the gradient to be orthogonal to the input features, which ensures the output features to change very slightly regarding the input.
To be specific, we maintain an orthogonal projection matrix set $\mP$ for the supernet. Each parameter in the supernet  has a corresponding projection matrix $\bP$, initialized as a unit matrix $\bI$. These  projection matrices are used to project gradients for model training.
Overall, there are two key issues:   how to update model parameters via the orthogonal projection matrix, and how to update the orthogonal projection matrix.

\subsubsection{The update of model parameters}
We train architectures based on orthogonal gradient descent~\cite{zeng2019continual}, \yifan{in which we project the gradient to be orthogonal to the input. In this way, we are able to update the model towards a useful direction but avoid large changes regarding the previous output of the shared parameters.} Formally,  given any architecture $\alpha_i$
and for any layer $\bW\in w(\alpha_i)$, there is a corresponding projection matrix $\bP$ inherited from the projection matrix set $\mP$. We update $\bW$ by the following orthogonal gradient descent:
\begin{align}\label{eq:updating_share}
\bW\leftarrow\bW-\eta\bP\nabla_{\bW}\mL(\alpha_i,w(\alpha_i)),
\end{align}
where $\bP\nabla_{\bW}\mL(\cdot)$ is often called  as the projected gradient.
Moreover, when the layers in $\alpha_i$ update, the corresponding layers in the supernet also update.

\subsubsection{The update of projection matrices}
In order to project the gradient of each shared layer to avoid large changes regarding its output, we ensure the projected gradient is orthogonal to the previous input. Specifically, for any layer $\bW$ of $\alpha_i$, let  $\bX\in\mmR^{d\times N}$ denote the input feature maps of all previously trained architectures, with a total of $N$ feature maps (whose dimension is $d$). We compute the orthogonal projection matrix $\bP$ by:
\begin{align}\label{eq:compute_p}
\bP =&~\bI-\bX(\lambda\bI + \bX^{\top}\bX)^{-1}\bX^{\top}, \end{align}
where $\lambda$ is some regularization constant.
The detailed derivations will be provided in a long version.
Based on Eqn.~(\ref{eq:compute_p}), our method is able to project the gradient to be orthogonal to the input, which helps to void a large change regarding the output of the shared layers (see Theorem~\ref{prof1} for more discussion).

A potential issue of Eqn.~(\ref{eq:compute_p}) is that the computation of $\bP$ requires all previous input feature maps (as described in Section~\ref{sec:motivation}). However, such a manner may be highly storage consuming as the input feature maps increase.
To handle this, we can update the projection matrix in an iterative manner~\cite{zhang2018online,zhang2019online,zhao2018adaptive}. For each coming sample $\bx\in\mmR^d$, based on Woodbury identity~\cite{horn2012matrix}, we update $\bP$ by:
\begin{align}\label{eq:update_p}
    \bP \leftarrow \bP - (\bP\bx \bx^{\top} \bP)/(\lambda + \bx^\top \bP \bx),
\end{align}
Based on Eqn.~(\ref{eq:update_p}), we update $\bP$ iteratively and each iteration only requires the input feature maps \wrt a single sample. By doing so, we need not to store all previous input feature maps, thus avoiding the high storage issue. Note that the update of $\bP$ also means the update of the corresponding projection matrix in $\mP$.

\subsection{Disturbance-immune Neural Architecture Search}

The proposed disturbance-immune update strategy helps to alleviate the issue of performance disturbance in the WS, and provides more stable performance estimations for candidate architectures. As a result, we can obtain more accurate reward signals for learning a good controller.
Following this, based on the disturbance-immune update strategy,
we propose a new disturbance-immune training scheme for neural architecture search (named as \irnas). The overall training scheme is provided in Algorithm~\ref{alg:overall}.

The main difference between the proposed \irnas  and other standard WS based NAS  is the training of the supernet. Specifically, we train the supernet via the new proposed disturbance-immune update strategy, which alleviates the performance disturbance issue in the training process. To be specific, we maintain an orthogonal projection matrix set $\mP$ for  the supernet (see line 2 in Algorithm~\ref{alg:overall}). Each parameter in the supernet has a corresponding projection matrix $\bP$, initialized as a unit matrix $\bI$. Based on the orthogonal projection matrix, we update the network architecture by orthogonal gradient descent (see lines 9-10 in Algorithm~\ref{alg:overall}).

\begin{algorithm}[tb]
	\caption{\small{Overall training scheme of \irnas}}\label{alg:overall}
    \begin{algorithmic}[1]\small
    \REQUIRE Training data $\mD_{train}$, validation data $\mD_{val}$, learning rate $\eta$, parameters $T_n$,$T_c$.
    \STATE {Initialize supernet parameters $w$ and controller parameters $\theta$.}
    \STATE Construct and initialize a projection matrix set $\mP$ for the supernet.\\
    \WHILE {not convergent}
    \STATE // \emph{Update $w$ by minimizing the training loss}
    \FOR{$i=1,...,T_n$}
    \STATE Sample $\alpha\sim\pi(\alpha;\theta)$.~~~\small{//} \emph{ $\pi(\cdot)$ is the policy of the controller}
    \STATE Sample a batch of data
    from $\mD_{train}$.
    \STATE \small{//} \emph{\textbf{Disturbance-immune Update} for $w(\alpha)$ and $\mP(\alpha)$}
    \STATE {Update $w(\alpha)$ using Eqn.~(\ref{eq:updating_share}).}~~~\small{//} \emph{$w(\alpha)$ denotes the parameters of $\alpha$}
    \STATE {Update $\mP(\alpha)$ using Eqn.~(\ref{eq:update_p}).}~~~\small{//} \emph{$\mP(\alpha)$ denotes the projection matrices of $\alpha$}
    \ENDFOR
    \STATE // \emph{Update $\theta$ by maximizing the reward}
    \FOR{$j=1,...,T_c$}
    \STATE Sample $\alpha\sim\pi(\alpha;\theta)$.
    \STATE Sample a batch of data  from $\mD_{val}$.
    \STATE Update
    $\theta\leftarrow\theta+\eta\mathcal{R}(\alpha, w)\nabla_\theta\text{log}\pi(\alpha;\theta)$.
    \ENDFOR
    \ENDWHILE
    \end{algorithmic}
	\label{alg:training}
    \vspace{-0.03in}
 \end{algorithm}

\subsection{Theoretical Analysis}\label{sec:theory}

In this section, we theoretically analyze the proposed method regarding its effectiveness and convergence.
To begin with, we analyze the effectiveness of our proposed method in alleviating the issue of performance disturbance as follows.

\begin{theorem}\label{prof1}
Given a model with parameter $\bW\small{\in}\mmR^{d\times m}$ and any input matrix  $\bX\small{\in}\mmR^{d\times N}$,  let $\bP=\bI-\bX(\lambda\bI+\bX^\top\bX)^{-1}\bX^\top$ be the projection matrix and $\bG$ be the gradient \wrt $\bW$. For the update of $\bW$ in the direction $\bP\bG$ and  $\eta\in\mmR^+$, let $\Delta(\bX)=||(\bW\small{-}\eta \bP\bG)^\top\bX\small{-}\bW^\top\bX ||_\mathrm{F}$ denote the output change of the original model and the updated model. When $\lambda\geq 0$, the following inequality holds:
\begin{align}\label{eq:theorem1}
    \Delta(\bX)
    \leq \eta\lambda\sqrt{d}  ||\bG^\top||_\mathrm{F}D(\bX),
\end{align}
where $D(\bX)=||\bX||_\mathrm{F}R(\bX)(1\small{+}\lambda R(\bX))$ and $R(\bX)=||(\bX\bX^\top)^{\small{-1}}||_2$.
\end{theorem}

Theorem~\ref{prof1} indicates that for any input feature map $\bX$, the distance between the original output $\bW^\top\bX$ and the one after the update $(\bW\small{-}\eta \bP\bG)^\top\bX$ is controlled by the regularization factor $\lambda$. Specifically, the lower $\lambda$, the lower distance upper bound.  Therefore, our proposed orthogonal gradient descent method is able to satisfy the constraint in Eqn.~(\ref{eq:new_constraint}). In other words, our method can avoid the large change regarding the previous output of shared layers when training the supernet, thus alleviating the issue of performance disturbance in WS.

We further provide an asymptotic analysis regarding the convergence of the update for $\bW$ as follows.

\begin{theorem} \label{theorem:convergence}
Given a loss function $\mathcal{L}(\bW)$ that is $L$-smooth and convex \emph{\wrt}~$\bW$. Let $\bW^*$ and $\bW_0$ be the optimal and initial solution of $\mL(\cdot)$. By setting  $\eta = 1/L$,  at the $t$-th update  step, the proposed update strategy  satisfies:
\begin{equation}\label{eq:convergence}
        \mL(\bW_t) - \mL(\bW^*) \leq\frac{2L}{t}||\bW_0-\bW^*||_\mathrm{F}^2.
\end{equation}
\end{theorem}

This theorem illustrates that our proposed update strategy has a sublinear convergence rate of $O(1/t)$, which guarantees the effectiveness of the proposed method.

\section{Experimental Results}
We evaluate the proposed \irnas in two main aspects: (1) the superiority of the searched architecture by \irnas on CIFAR-10 and ImageNet, respectively; (2) the effectiveness of our proposed disturbance-immune update strategy. The source code will be publicly available.


\subsection{Evaluation on CIFAR-10}
In this section, we evaluate the proposed \irnas method on CIFAR-10~\cite{krizhevsky2009learning}. To be specific, we first use the proposed method to train a controller on CIFAR-10, and use it to search for a convolutional neural architecture. By comparing the searched architecture with other state-of-the-art architectures, we can evaluate the effectiveness of our method.  To this end, we first describe the search space, training details and evaluation details.

\textbf{Search space.} Following the settings in DARTS~\cite{liu2018darts}, we aim to search for two types of convolutional cells, namely the normal cell and the reduction cell. Each cell contains 7 nodes, including 2 input nodes, 4 intermediate nodes and 1 output node. Between any two nodes, there are 8 available operations, including $3\small{\times} 3$ depthwise separable convolution, $3\small{\times} 3$ dilated convolution, $3\small{\times} 3$ max pooling, $3\small{\times} 3$ average pooling, $5\small{\times} 5$ depthwise separable convolution, $5\small{\times} 5$ dilated convolution, identity and none. After obtaining the convolutional cells, we stack them to build the final convolutional network.

\textbf{Training details.}
In the search phase, we divide the standard training set of CIFAR-10 into two parts.  Specifically, we randomly select 40\% of the training set  to train sampled architectures (as training data) and use the rest 60\% to learn controllers (as validation data). Moreover, we train \irnas for $240$ epochs in total. We first train the supernet without learning the controller, and start training the controller from epoch 90. In addition, we set $\lambda=1$ by default, where the sensitivity analysis can be found in Section~\ref{sec_discussions}. For training the supernet, we use an SGD optimizer with a weight decay of $3 \times 10^{-4}$ and a momentum of $0.9$. The learning rate is set to $0.1$.
For training the controller, we use ADAM with a learning rate of $3 \times 10^{-4}$ and a weight decay of $5 \times 10^{-4}$.
We add the controller's sample entropy to the reward, which is weighted by $0.005$.

\textbf{Evaluation details.}
In the evaluation phase, we first use the learned controller to search for a normal cell and a reduction cell. Then, we construct the final convolutional network with 17 normal cells and 2 reduction cells.  Following~\cite{nayman2019xnas}, we put the two reduction cells at the $1/3$ and $2/3$ depth of the network, respectively. The initial number of the channels is set to 43. Following DARTS~\cite{liu2018darts}, we train the convolutional network for $600$ epochs with a batch size of $128$. We apply \ljc{an} SGD optimizer with a weight decay of $3 \times 10^{-4}$ and a momentum of $0.9$. Moreover, we set the initial learning rate as $0.05$ and use the cosine annealing strategy~\cite{SGDR} to adjust it. We also use the cutout scheme~\cite{devries2017improved} with length 16 for data augmentation.

	\begin{table*}[t]
	\vspace{0.15in}
		\centering
		\caption{Comparisons with state-of-the-art NAS methods on CIFAR-10. Moreover, ``-" means unavailable results.
		}
		\label{tab:cifar-10}
        {

    \begin{tabular*}{0.95\textwidth}{@{}@{\extracolsep{\fill}}cccccc@{}}
	\hline
	\multicolumn{1}{c}{\multirow{2}[0]{*}{Architecture}} &
	\multicolumn{1}{c}{\multirow{2}[0]{*}{Test Accuracy (\%)}} & &
	\multicolumn{1}{c}{\multirow{2}[0]{*}{\# Params (M)}} & &
    \multicolumn{1}{c}{Search Cost} \\
    &  & &  & & (GPU days)  \\
			\hline
			DenseNet-BC~\cite{huang2017densely}&96.54& &25.6 & & -- \\
			PyramidNet-BC~\cite{han2017deep}&96.69& &26.0 & & -- \\
			\hline
			Random search baseline &96.71 $\pm$ 0.15 & &3.2 & & -- \\
			NASNet-A + cutout~\cite{zoph2018learning}&97.35 & &3.3 & & 1,800 \\
			NASNet-B~\cite{zoph2018learning}&96.27& &2.6 & & 1,800 \\
			NASNet-C~\cite{zoph2018learning}&96.41& &3.1 & & 1,800 \\
			AmoebaNet-A + cutout~\cite{real2018regularized}&96.66 $\pm$ 0.06& &3.2 & & 3,150\\
			AmoebaNet-B + cutout~\cite{real2018regularized}&96.63 $\pm$ 0.04& &2.8 & & 3,150\\
			Hierarchical Evo~\cite{liu2017hierarchical}&96.25 $\pm$ 0.12 & &15.7 & & 300 \\
			SNAS~\cite{xie2018snas}&97.02& &2.9 & & 1.5\\
			GHN~\cite{zhang2018graph} & 97.16 $\pm$ 0.07 & & 5.7 & & 0.8 \\
			ENAS + cutout~\cite{pham2018efficient}&97.11& &4.6 & & 0.5 \\
			DARTS + cutout~\cite{liu2018darts} &97.24 $\pm$ 0.09 & &3.4& & 4\\
			NAT-DARTS~\cite{NAT} &97.28 & &2.7 & & --\\
			NAONet~\cite{luo2018neural}&97.02& &28.6 & & 200\\
			NAONet-WS~\cite{luo2018neural}&96.47& &2.5 & & 0.3\\
			\hline
			\irnas + cutout & \textbf{97.38 $\pm$ 0.04} & & 3.7 & & 1.5 \\
			\hline		
		\end{tabular*}
		}
	\end{table*}
	
\begin{figure*}[!h]
\centering
\subfigure[Normal cell.]
{
	\label{fig:normal}
	\includegraphics[height=4.5cm]{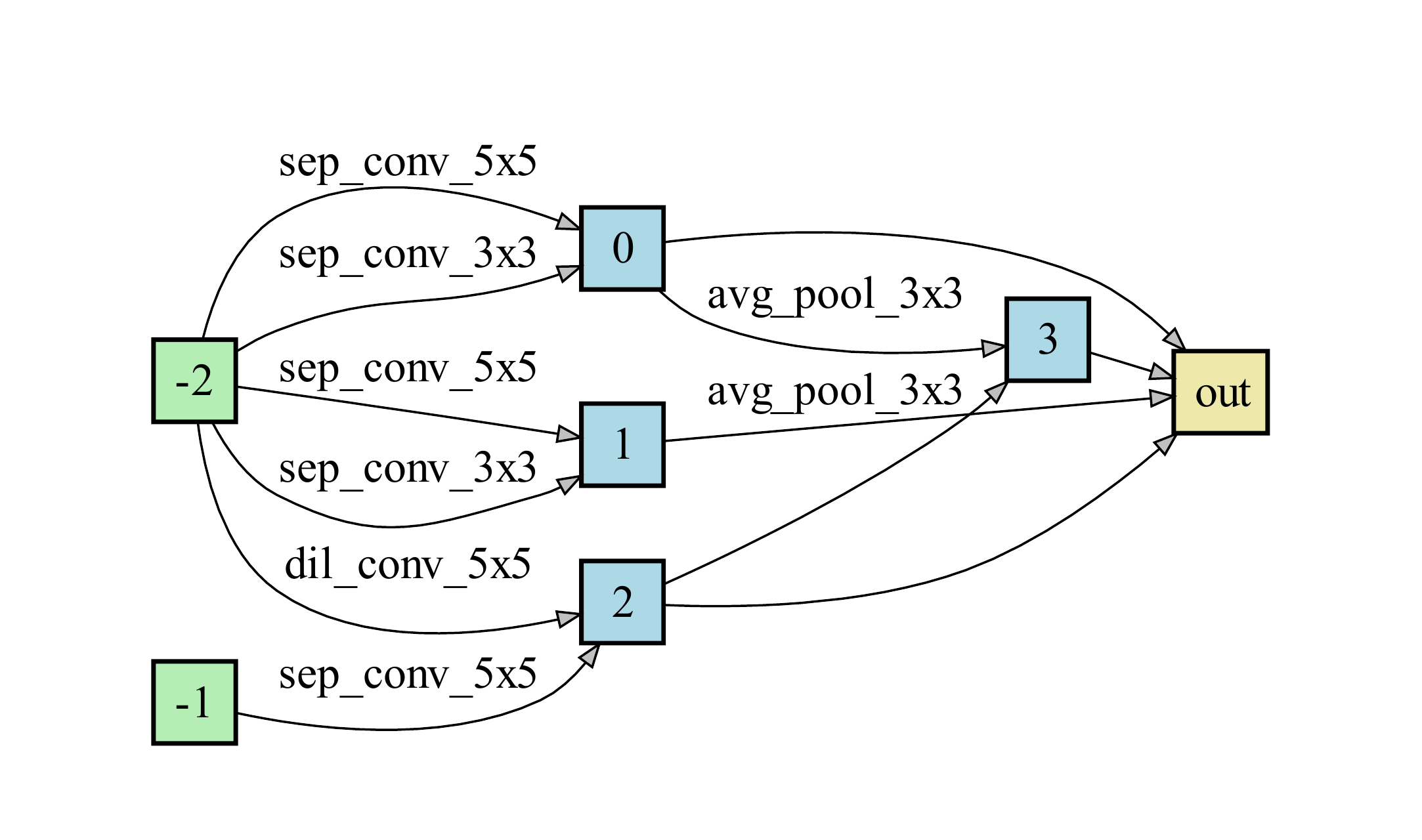}
}
\subfigure[Reduction cell.]
{
	\label{fig:reduction}
	\includegraphics[height = 5.2cm]{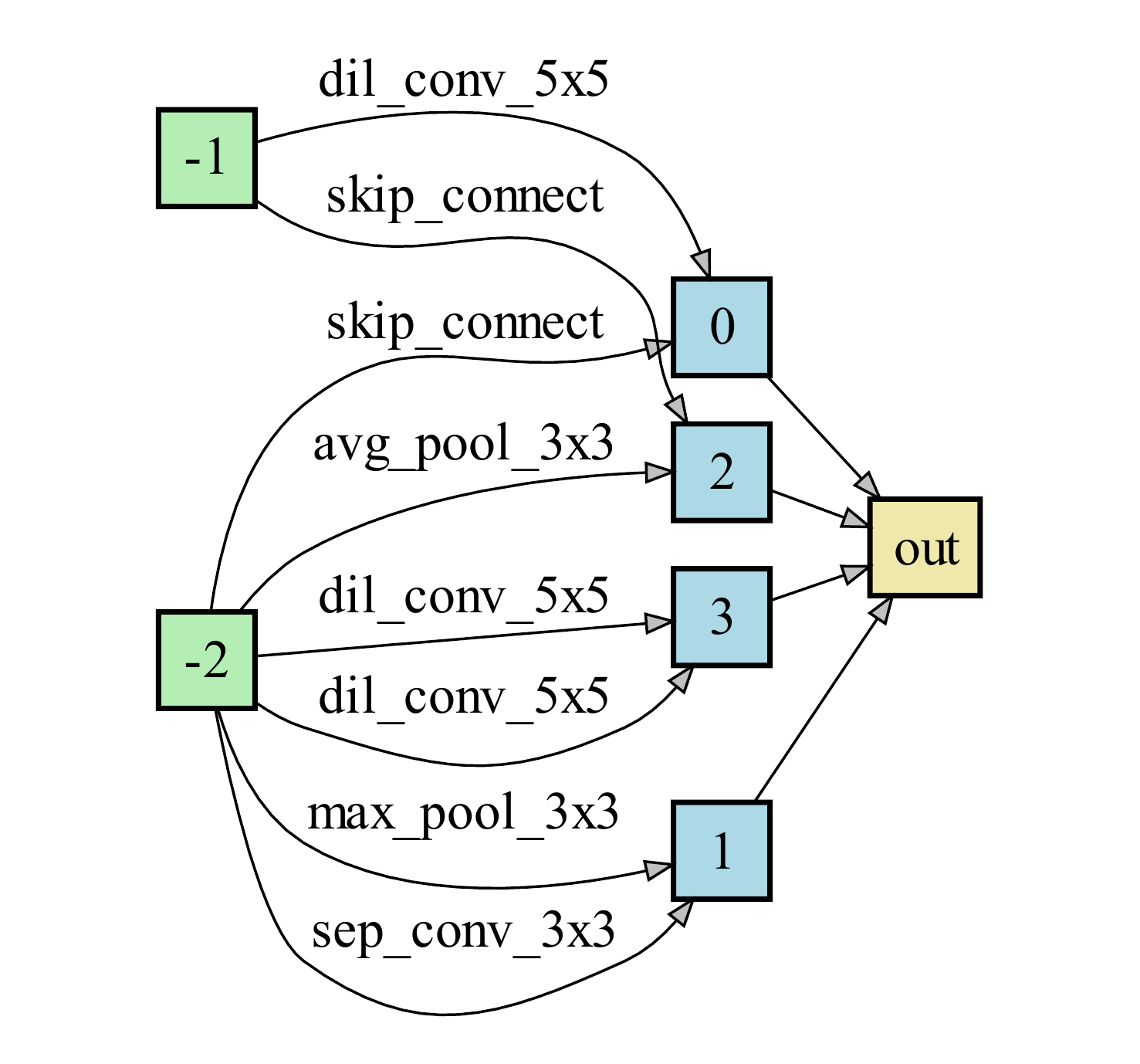}
}
\caption{The searched convolutional cells by \irnas on the CIFAR-10 dataset.}
\label{fig:cnn_arch}
\end{figure*}

\textbf{Comparison with state-of-the-art methods.}
We show the searched normal cells and reduction cells in Fig.~\ref{fig:cnn_arch} and report the detailed results of all methods in Table~\ref{tab:cifar-10}. For our searched architectures, we run 5 experiments with the different random initialization, and report the average performance with standard deviation. Experimental results show that the architecture searched by our \irnas achieves 97.38\% accuracy, outperforming all other state-of-the-art architectures, including human-designed ones and NAS ones. Note that ENAS, NAONet-WS, DARTS, and NAT also use the scheme of weight sharing (WS).
Such a result demonstrates that it is necessary to alleviate the issue of performance disturbance(PD) in WS-based NAS. This helps us to  provide more accurate rewards for controller learning, and search for better neural architectures.

\subsection{Evaluation on ImageNet}
To verify the generalization of the  convolutional cells searched on CIFAR-10 (as shown in Fig.~\ref{fig:cnn_arch}), we further evaluate them on a large-scale image classification dataset, namely ImageNet~\cite{deng2009imagenet}.
To begin with, we describe the evaluation details.
	
\textbf{Evaluation details.}
For the ImageNet dataset, we construct the convolutional network with 12 normal cells and 2 reduction cells. We put the two reduction cells at the $1/3$ and $2/3$ depth of the network, respectively. We set the number of the initial channels to 48. Following~\cite{liu2018darts}, we train the network by \ljc{an} SGD optimizer with $250$ epochs and use a weight decay of $3 \times 10^{-5}$ and a momentum of $0.9$. We initialize the learning rate as $0.1$ and decrease it by the cosine annealing~\cite{SGDR}. We follow the ImageNet mobile setting~\cite{liu2018darts}, where the size of input images is set to $224 \times 224$ and the number of multiply-adds (Madds) is less than 600M.

\begin{table*}[t]
	\centering
	\caption{Comparison results on ImageNet, where we use the evaluation code of DARTS~\cite{liu2018darts}. Moreover, ``-'' means unavailable results.
	}
	\label{tab:imagenet}
	{
    \begin{tabular*}{0.99\textwidth}{@{}@{\extracolsep{\fill}}ccccccccc@{}}
    \hline
    \multicolumn{1}{c}{\multirow{2}[0]{*}{Architecture}} &
    \multicolumn{2}{c}{Test Accuracy (\%)} & &
    \multicolumn{1}{c}{\# Params} & &
    \multicolumn{1}{c}{\# MAdds} & &
    \multicolumn{1}{c}{Search Cost} \\
    \cline{2-3}
    & \multicolumn{1}{c}{~~Top-1} & \multicolumn{1}{c}{~~Top-5} &  & (M)    &   & (M)  & & (GPU days)  \\
    \hline
    ResNet-18~\cite{resnet} & ~~69.8 & ~~89.1 & & 11.7 & & 1,814 & &--\\
    Inception-v1~\cite{szegedy2015going} & ~~69.8      & ~~89.9    &  &   6.6  &  &  1,448 & &--\\
    MobileNet v1 ($1\times$)~\cite{howard2017mobilenets} & ~~70.6 & ~~89.5 & & 4.2 & & 569 & &-- \\
    ShuffleNet v1 ($2\times$)~\cite{Zhang2018ShuffleNetAE} & ~~70.9 & ~~89.2 & & 5.0 & & 524 & &--\\
    \hline
    NASNet-A~\cite{zoph2018learning} & ~~74.0 & ~~91.6 & & 5.3 & & 564 & & 3,150\\
    NASNet-B~\cite{zoph2018learning} & ~~72.8 & ~~91.3 & & 5.3 & & 488 & & 3,150\\
    NASNet-C~\cite{zoph2018learning} & ~~72.5 & ~~91.0 & & 4.9 & & 558 & & 3,150\\
    AmoebaNet-A~\cite{real2018regularized} & ~~74.5 & ~~92.0 & & 5.1 & & 555 & & 1,800\\
    AmoebaNet-B~\cite{real2018regularized} & ~~74.0 & ~~91.5 & & 5.3 & & 555 & & 1,800\\
    GHN~\cite{zhang2018graph} & ~~73.0 & ~~91.3 & & 6.1 & & 569 & & 0.8\\
    PNAS~\cite{liu2018progressive} & ~~74.2 & ~~91.9 & & 5.1 & & 588 & & 255\\
    BayesNAS~\cite{BayesNAS} & ~~73.5 & ~~91.1 & & 3.9 & & -  & & 0.2\\
    DARTS~\cite{liu2018darts} & ~~73.1 & ~~91.0 & & 4.9 & & 595 & & 4\\
    NAT-DARTS~\cite{NAT} & ~~73.7 & ~~91.4 & & 4.0 & & 441 & & --\\
    SNAS~\cite{xie2018snas} & ~~72.7 & ~~90.8 & & 4.3 & & 522  & & 1.5\\
    \hline
    \irnas & ~~\textbf{74.7} & ~~\textbf{92.1} & & 5.2 & & 587 & & 1.5\\
    \hline
    \end{tabular*}
    }\vspace{-0.15in}
\end{table*}

\textbf{Comparison with state-of-the-art methods.}
We compare our searched architecture with several state-of-the-art models on ImageNet. As shown in Table~\ref{tab:imagenet}, our architecture achieves 74.7\% top-1 accuracy and 92.1\% top-5 accuracy. To be specific,  our architecture outperforms human-designed architectures (\eg{ResNet-18~\cite{resnet}}) by about 5\%, and outperforms most of NAS models by $0.5\% \backsim 2\%$ in terms of {top-1 accuracy}. Moreover, our architecture achieves excellent performance only using 1.5 GPU days, while AmoebaNet~\cite{real2018regularized} and NASNet~\cite{zoph2018learning} spend 3,150 and 1,800 GPU days, respectively. These results  demonstrate the generalization of the searched convolutional cells and the effectiveness/efficiency of the proposed method.

\subsection{Effectiveness of Disturbance-immune Update Strategy}
In previous experiments, we have demonstrated the superiority of the proposed method. One important reason for superiority is the ability to alleviate performance disturbance (PD). In this section, we further verify the effectiveness of our proposed update strategy in dealing with PD. To this end, we use the following two metrics to measure the degree of PD and use them to evaluate the proposed method.

\textbf{\ljc{Metrics for PD.}}
(1) Performance change: for any architecture $\alpha$ inherited from the supernet, performance change is defined as ($acc_j\small{-}acc_i$), where $acc_j$ and $acc_i$ denote the validation accuracy of $\alpha$ at the $j$-th and $i$-th training epoch of the supernet\ljc{, respectively}. Note that $j>i$ and  $(j\small{-}i)$ indicates the \textbf{epoch interval}. Overall, the larger performance change indicates the more severe performance disturbance.
(2) Kendalls Tau (KTau): we use KTau~\cite{sen1968estimates} to measure the correlation of performance ranks between two architecture sets. The range of KTau belongs to $[-1,1]$, where larger KTau means that the performance ranks of two architecture sets are more consistent.

\begin{figure*}[ht]
\centering
\begin{subfigure}[absolute performance change]{
\centering
\includegraphics[width=6.4cm]{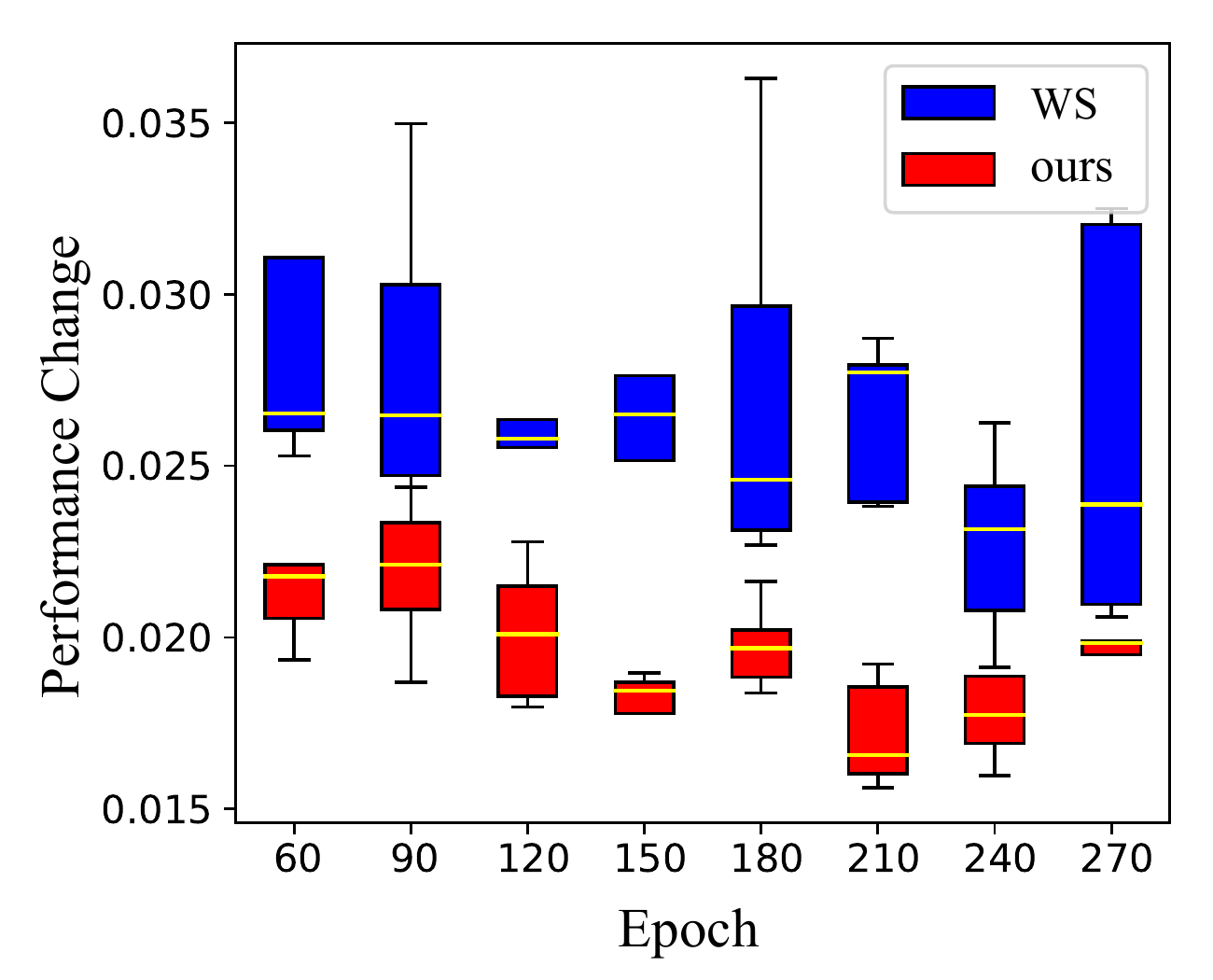}}
\end{subfigure}
\begin{subfigure}[KTau at different training epochs]{
\centering
\includegraphics[width=6.4cm]{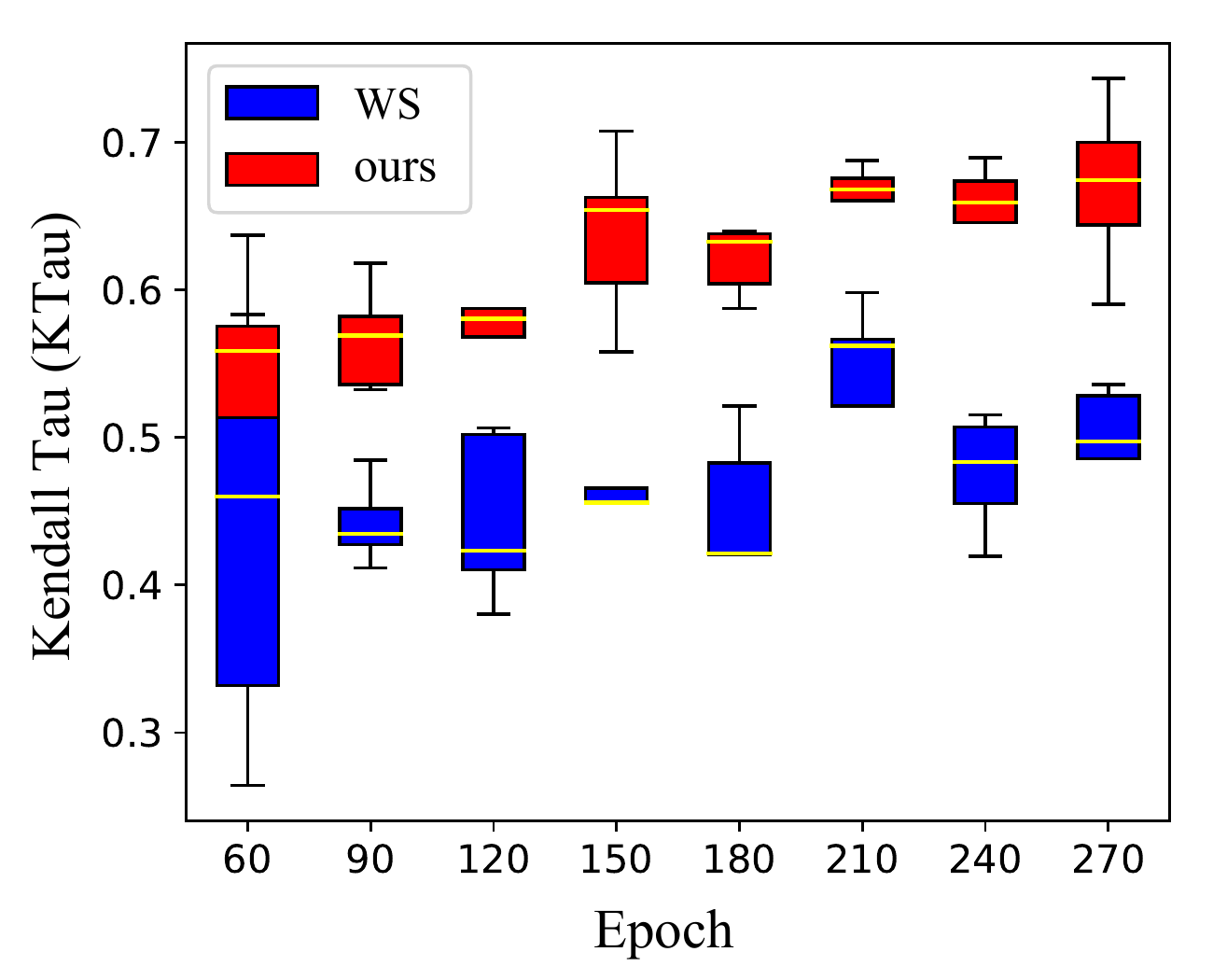}}
\end{subfigure}
\vspace{-0.1in}
\caption{Comparisons between our proposed disturbance-immune weight sharing (WS) scheme and the standard WS scheme. The larger the performance change is, the more severe the PD; the higher KTau is, the more accurate the performance estimation.
}
\label{fig:hpnas_epoch_interval_13}
\end{figure*}

\textbf{Evaluation in terms of performance change.}
To evaluate our method, we randomly sample 64 architectures based on the supernet. We record their performance at different training epochs of the supernet, and report the average performance change of these 64 architectures regarding epoch interval 13.
Fig.~\ref{fig:hpnas_epoch_interval_13}~(a) shows that our disturbance-immune WS scheme is able to reduce the performance change of architectures and thus alleviate the PD issue.

\textbf{Evaluation in terms of KTau.}
Based on the above architectures, we compute the KTau between their current performance rank and the one after 13 epochs (\ie epoch interval is 13). Fig.~\ref{fig:hpnas_epoch_interval_13} (b) verifies the effectiveness of our method in improving KTau. Note that the higher KTau indicates a more consistent performance estimation in the training process, which means more stable/accurate rewards for controller learning. Moreover, we compute the ground-truth KTau (GT-Tau)~\cite{sciuto2019evaluating}, which measures the performance correlation between a set of architectures inherited from the supernet and exactly trained from scratch. Specificcally, our method achieves higher GT-Tau (0.48) than the standard WS (0.16).

\subsection{More Discussions}\label{sec_discussions}
We further discuss PD for the different numbers of shared layers and parameter sensitivity $\wrt \lambda$ through some self-designed experiments.
We provide our main observations \ljc{and analyses} as follows.

\textbf{Number of shared layers.}
We find that as the number of shared layers \ljc{between two} architectures increases, the issue of PD becomes more severe.
Since there often exist many shared layers in WS-based NAS, this result demonstrates that it is necessary to alleviate the issue of PD for NAS.

\textbf{Regularization parameter $\lambda$.}
Generally, the optimal value of $\lambda$ varies regarding different data.
A large $\lambda$ may make the method fail to satisfy the problem constraint (see Eqn.~(\ref{eq:new_constraint})), while a small $\lambda$ may result in the irreversible issue when computing projection matrices (see Eqn.~(\ref{eq:compute_p})).
Nevertheless, the default setting $\lambda=1$ helps to achieve the best or relatively good performance in most cases.

%
%
%

\section{Conclusions}

In this paper, we have proposed a novel disturbance-immune training scheme for NAS to conquer performance disturbance (PD) in weight sharing (WS). Specifically, by developing a new update strategy to train sampled architectures, our method provides more stable/accurate performance estimation for architectures. As a result, the proposed method is able to learn a good controller for  searching good architectures.
We theoretically and empirically verify the effectiveness of the proposed method in alleviating PD, and also provide an asymptotic analysis for its convergence. Extensive experiments demonstrate that the architecture found by our proposed method outperforms the architectures obtained by considered state-of-the-art NAS methods.

\clearpage

\bibliographystyle{abbrv}
{
	\small
	\bibliography{reference}

\begin{thebibliography}{10}

\bibitem{adam2019understanding}
G.~Adam and J.~Lorraine.
\newblock Understanding neural architecture search techniques.
\newblock {\em arXiv}, 2019.

\bibitem{baker2016designing}
B.~Baker, O.~Gupta, N.~Naik, and R.~Raskar.
\newblock Designing neural network architectures using reinforcement learning.
\newblock In {\em International Conference on Learning Representations}, 2017.

\bibitem{bender2018understanding}
G.~Bender, P.-J. Kindermans, B.~Zoph, V.~Vasudevan, and Q.~Le.
\newblock Understanding and simplifying one-shot architecture search.
\newblock In {\em International Conference on Machine Learning}, pages
  549--558, 2018.

\bibitem{cai2018efficient}
H.~Cai, T.~Chen, W.~Zhang, Y.~Yu, and J.~Wang.
\newblock Efficient architecture search by network transformation.
\newblock In {\em AAAI Conference on Artificial Intelligence}, pages
  2787--2794, 2018.

\bibitem{cai2018proxylessnas}
H.~Cai, L.~Zhu, and S.~Han.
\newblock Proxylessnas: Direct neural architecture search on target task and
  hardware.
\newblock In {\em International Conference on Learning Representations}, 2019.

\bibitem{cao2019multi}
J.~Cao, L.~Mo, Y.~Zhang, et~al.
\newblock Multi-marginal wasserstein gan.
\newblock In {\em Advances in Neural Information Processing Systems}, pages
  1774--1784, 2019.

\bibitem{chen2019detnas}
Y.~Chen, T.~Yang, X.~Zhang, G.~Meng, C.~Pan, and J.~Sun.
\newblock Detnas: Neural architecture search on object detection.
\newblock {\em arXiv}, 2019.

\bibitem{chu2019fairnas}
X.~Chu, B.~Zhang, R.~Xu, and J.~Li.
\newblock Fairnas: Rethinking evaluation fairness of weight sharing neural
  architecture search.
\newblock {\em arXiv}, 2019.

\bibitem{deng2009imagenet}
J.~Deng, W.~Dong, R.~Socher, L.-J. Li, K.~Li, and L.~Fei-Fei.
\newblock Imagenet: A large-scale hierarchical image database.
\newblock In {\em IEEE Conference on Computer Vision and Pattern Recognition},
  pages 248--255, 2009.

\bibitem{devries2017improved}
T.~DeVries and G.~W. Taylor.
\newblock Improved regularization of convolutional neural networks with cutout.
\newblock {\em arXiv}, 2017.

\bibitem{farajtabar2019orthogonal}
M.~Farajtabar, N.~Azizan, A.~Mott, and A.~Li.
\newblock Orthogonal gradient descent for continual learning.
\newblock {\em arXiv}, 2019.

\bibitem{finn2017model}
C.~Finn, P.~Abbeel, and S.~Levine.
\newblock Model-agnostic meta-learning for fast adaptation of deep networks.
\newblock In {\em International Conference on Machine Learning}, pages
  1126--1135, 2017.

\bibitem{ghiasi2019fpn}
G.~Ghiasi, T.-Y. Lin, and Q.~V. Le.
\newblock Nas-fpn: Learning scalable feature pyramid architecture for object
  detection.
\newblock In {\em IEEE Conference on Computer Vision and Pattern Recognition},
  pages 7036--7045, 2019.

\bibitem{guo2018double}
Y.~Guo, Q.~Wu, C.~Deng, J.~Chen, and M.~Tan.
\newblock Double forward propagation for memorized batch normalization.
\newblock In {\em Thirty-Second AAAI Conference on Artificial Intelligence},
  2018.

\bibitem{NAT}
Y.~Guo, Y.~Zheng, M.~Tan, Q.~Chen, J.~Chen, P.~Zhao, and J.~Huang.
\newblock {NAT:} neural architecture transformer for accurate and compact
  architectures.
\newblock In {\em Advances in Neural Information Processing Systems}, pages
  735--747, 2019.

\bibitem{han2017deep}
D.~Han, J.~Kim, and J.~Kim.
\newblock Deep pyramidal residual networks.
\newblock In {\em IEEE Conference on Computer Vision and Pattern Recognition},
  pages 6307--6315, 2017.

\bibitem{resnet}
K.~He, X.~Zhang, S.~Ren, and J.~Sun.
\newblock Deep residual learning for image recognition.
\newblock In {\em IEEE Conference on Computer Vision and Pattern Recognition},
  pages 770--778, 2016.

\bibitem{he2017overcoming}
X.~He and H.~Jaeger.
\newblock Overcoming catastrophic interference by conceptors.
\newblock {\em arXiv}, 2017.

\bibitem{horn2012matrix}
R.~A. Horn and C.~R. Johnson.
\newblock {\em Matrix analysis}.
\newblock Cambridge university press, 2012.

\bibitem{howard2017mobilenets}
A.~G. Howard, M.~Zhu, B.~Chen, D.~Kalenichenko, W.~Wang, T.~Weyand,
  M.~Andreetto, and H.~Adam.
\newblock Mobilenets: Efficient convolutional neural networks for mobile vision
  applications.
\newblock {\em arXiv}, 2017.

\bibitem{huang2017densely}
G.~Huang, Z.~Liu, L.~Van Der~Maaten, and K.~Q. Weinberger.
\newblock Densely connected convolutional networks.
\newblock In {\em IEEE Conference on Computer Vision and Pattern Recognition},
  pages 2261--2269, 2017.

\bibitem{kirkpatrick2017overcoming}
J.~Kirkpatrick, R.~Pascanu, N.~Rabinowitz, J.~Veness, G.~Desjardins, A.~A.
  Rusu, K.~Milan, J.~Quan, T.~Ramalho, A.~Grabska-Barwinska, et~al.
\newblock Overcoming catastrophic forgetting in neural networks.
\newblock {\em National Academy of Sciences}, pages 3521--3526, 2017.

\bibitem{krizhevsky2009learning}
A.~Krizhevsky, G.~Hinton, et~al.
\newblock Learning multiple layers of features from tiny images.
\newblock 2009.

\bibitem{li2019random}
L.~Li and A.~Talwalkar.
\newblock Random search and reproducibility for neural architecture search.
\newblock In {\em Conference on Uncertainty in Artificial Intelligence}, 2019.

\bibitem{liu2019auto}
C.~Liu, L.-C. Chen, F.~Schroff, H.~Adam, W.~Hua, A.~L. Yuille, and L.~Fei-Fei.
\newblock Auto-deeplab: Hierarchical neural architecture search for semantic
  image segmentation.
\newblock In {\em IEEE Conference on Computer Vision and Pattern Recognition},
  pages 82--92, 2019.

\bibitem{liu2018progressive}
C.~Liu, B.~Zoph, M.~Neumann, J.~Shlens, W.~Hua, L.-J. Li, L.~Fei-Fei,
  A.~Yuille, J.~Huang, and K.~Murphy.
\newblock Progressive neural architecture search.
\newblock In {\em European Conference on Computer Vision}, pages 19--34, 2018.

\bibitem{liu2017hierarchical}
H.~Liu, K.~Simonyan, O.~Vinyals, C.~Fernando, and K.~Kavukcuoglu.
\newblock Hierarchical representations for efficient architecture search.
\newblock In {\em International Conference on Learning Representations}, 2017.

\bibitem{liu2018darts}
H.~Liu, K.~Simonyan, and Y.~Yang.
\newblock Darts: Differentiable architecture search.
\newblock In {\em International Conference on Learning Representations}, 2019.

\bibitem{SphereFace}
W.~Liu, Y.~Wen, Z.~Yu, M.~Li, B.~Raj, and L.~Song.
\newblock Sphereface: Deep hypersphere embedding for face recognition.
\newblock In {\em IEEE Conference on Computer Vision and Pattern Recognition},
  pages 6738--6746, 2017.

\bibitem{SGDR}
I.~Loshchilov and F.~Hutter.
\newblock {SGDR:} stochastic gradient descent with warm restarts.
\newblock In {\em International Conference on Learning Representations}, 2017.

\bibitem{luo2017thinet}
J.-H. Luo, J.~Wu, and W.~Lin.
\newblock Thinet: A filter level pruning method for deep neural network
  compression.
\newblock In {\em IEEE International Conference on Computer Vision}, pages
  5058--5066, 2017.

\bibitem{luo2019understanding}
R.~Luo, T.~Qin, and E.~Chen.
\newblock Understanding and improving one-shot neural architecture
  optimization.
\newblock {\em arXiv}, 2019.

\bibitem{luo2018neural}
R.~Luo, F.~Tian, T.~Qin, E.~Chen, and T.-Y. Liu.
\newblock Neural architecture optimization.
\newblock In {\em Advances in Neural Information Processing Systems}, pages
  7816--7827, 2018.

\bibitem{nayman2019xnas}
N.~Nayman, A.~Noy, T.~Ridnik, I.~Friedman, R.~Jin, and L.~Zelnik.
\newblock Xnas: Neural architecture search with expert advice.
\newblock In {\em Advances in Neural Information Processing Systems}, pages
  1975--1985, 2019.

\bibitem{pham2018efficient}
H.~Pham, M.~Y. Guan, B.~Zoph, Q.~V. Le, and J.~Dean.
\newblock Efficient neural architecture search via parameter sharing.
\newblock In {\em International Conference on Machine Learning}, pages
  4092--4101, 2018.

\bibitem{real2018regularized}
E.~Real, A.~Aggarwal, Y.~Huang, and Q.~V. Le.
\newblock Regularized evolution for image classifier architecture search.
\newblock In {\em AAAI Conference on Artificial Intelligence}, pages
  4780--4789, 2019.

\bibitem{schroff2015facenet}
F.~Schroff, D.~Kalenichenko, and J.~Philbin.
\newblock {F}acenet: {A} {U}nified {E}mbedding for {F}ace {R}ecognition and
  {C}lustering.
\newblock In {\em IEEE Conference on Computer Vision and Pattern Recognition},
  pages 815--823, 2015.

\bibitem{sciuto2019evaluating}
C.~Sciuto, K.~Yu, M.~Jaggi, C.~Musat, and M.~Salzmann.
\newblock Evaluating the search phase of neural architecture search.
\newblock In {\em International Conference on Learning Representations}, 2019.

\bibitem{sen1968estimates}
P.~K. Sen.
\newblock Estimates of the regression coefficient based on kendall's tau.
\newblock {\em Journal of American Statistical Association},
  63(324):1379--1389, 1968.

\bibitem{szegedy2015going}
C.~Szegedy, W.~Liu, Y.~Jia, P.~Sermanet, S.~Reed, D.~Anguelov, D.~Erhan,
  V.~Vanhoucke, and A.~Rabinovich.
\newblock Going deeper with convolutions.
\newblock In {\em IEEE Conference on Computer Vision and Pattern Recognition},
  pages 1--9, 2015.

\bibitem{tan2019mnasnet}
M.~Tan, B.~Chen, R.~Pang, V.~Vasudevan, M.~Sandler, A.~Howard, and Q.~V. Le.
\newblock Mnasnet: Platform-aware neural architecture search for mobile.
\newblock In {\em IEEE Conference on Computer Vision and Pattern Recognition},
  pages 2820--2828, 2019.

\bibitem{wu2019fbnet}
B.~Wu, X.~Dai, P.~Zhang, Y.~Wang, F.~Sun, Y.~Wu, Y.~Tian, P.~Vajda, Y.~Jia, and
  K.~Keutzer.
\newblock Fbnet: Hardware-aware efficient convnet design via differentiable
  neural architecture search.
\newblock In {\em IEEE Conference on Computer Vision and Pattern Recognition},
  pages 10734--10742, 2019.

\bibitem{xie2018snas}
S.~Xie, H.~Zheng, C.~Liu, and L.~Lin.
\newblock Snas: stochastic neural architecture search.
\newblock In {\em International Conference on Learning Representations}, 2019.

\bibitem{zeng2019continual}
G.~Zeng, Y.~Chen, B.~Cui, and S.~Yu.
\newblock Continual learning of context-dependent processing in neural
  networks.
\newblock {\em Nature Machine Intelligence}, 1(8):364--372, 2019.

\bibitem{zhang2018graph}
C.~Zhang, M.~Ren, and R.~Urtasun.
\newblock Graph hypernetworks for neural architecture search.
\newblock In {\em International Conference on Learning Representations}, 2018.

\bibitem{Zhang2018ShuffleNetAE}
X.~Zhang, X.~Zhou, M.~Lin, and J.~Sun.
\newblock Shufflenet: An extremely efficient convolutional neural network for
  mobile devices.
\newblock In {\em IEEE Conference on Computer Vision and Pattern Recognition},
  pages 6848--6856, 2018.

\bibitem{zhang2019whole}
Y.~Zhang, H.~Chen, Y.~Wei, P.~Zhao, J.~Cao, et~al.
\newblock From whole slide imaging to microscopy: Deep microscopy adaptation
  network for histopathology cancer image classification.
\newblock In {\em International Conference on Medical Image Computing and
  Computer-Assisted Intervention}, pages 360--368, 2019.

\bibitem{zhang2019collaborative}
Y.~Zhang, Y.~Wei, P.~Zhao, S.~Niu, et~al.
\newblock Collaborative unsupervised domain adaptation for medical image
  diagnosis.
\newblock In {\em Medical Imaging meets NeurIPS}, 2019.

\bibitem{zhang2018online}
Y.~Zhang, P.~Zhao, J.~Cao, W.~Ma, et~al.
\newblock Online adaptive asymmetric active learning for budgeted imbalanced
  data.
\newblock In {\em Proceedings of the 24th ACM SIGKDD International Conference
  on Knowledge Discovery \& Data Mining}, pages 2768--2777. ACM, 2018.

\bibitem{zhang2019online}
Y.~Zhang, P.~Zhao, S.~Niu, Q.~Wu, et~al.
\newblock Online adaptive asymmetric active learning with limited budgets.
\newblock {\em IEEE Transactions on Knowledge and Data Engineering}, 2019.

\bibitem{zhang2018adaptive}
Y.~Zhang, P.~Zhao, Q.~Wu, B.~Li, et~al.
\newblock Cost-sensitive portfolio selection via deep reinforcement learning.
\newblock {\em IEEE Transactions on Knowledge and Data Engineering}, 2020.

\bibitem{zhao2018adaptive}
P.~Zhao, Y.~Zhang, M.~Wu, S.~C. Hoi, M.~Tan, and J.~Huang.
\newblock Adaptive cost-sensitive online classification.
\newblock {\em IEEE Transactions on Knowledge and Data Engineering},
  31(2):214--228, 2018.

\bibitem{BayesNAS}
H.~Zhou, M.~Yang, J.~Wang, and W.~Pan.
\newblock Bayesnas: {A} bayesian approach for neural architecture search.
\newblock In {\em International Conference on Machine Learning}, pages
  7603--7613, 2019.

\bibitem{zhuang2018discrimination}
Z.~Zhuang, M.~Tan, B.~Zhuang, J.~Liu, Y.~Guo, Q.~Wu, J.~Huang, and J.~Zhu.
\newblock Discrimination-aware channel pruning for deep neural networks.
\newblock In {\em Advances in Neural Information Processing Systems}, pages
  875--886, 2018.

\bibitem{zoph2016neural}
B.~Zoph and Q.~V. Le.
\newblock Neural architecture search with reinforcement learning.
\newblock In {\em International Conference on Learning Representations}, 2017.

\bibitem{zoph2018learning}
B.~Zoph, V.~Vasudevan, J.~Shlens, and Q.~V. Le.
\newblock Learning transferable architectures for scalable image recognition.
\newblock In {\em IEEE Conference on Computer Vision and Pattern Recognition},
  pages 8697--8710, 2018.

\end{thebibliography}
}

\end{document}